%% file: root.tex
\documentclass[letterpaper, 10 pt, conference]{ieeeconf}
\IEEEoverridecommandlockouts
\usepackage[utf8]{inputenc} 
\usepackage[english]{babel}

\usepackage[lined,ruled]{algorithm2e}
\usepackage{amsmath, amssymb, amsfonts}
\usepackage{bm}
\usepackage{booktabs}
\usepackage{cite}  
\usepackage{color}
\usepackage{graphicx}
\usepackage{siunitx}
\usepackage{soul} 
\usepackage{url}
\usepackage{hyperref}
\usepackage[dvipsnames]{xcolor}

\graphicspath{{./figures/}}

\title{\LARGE \bf 
SwarmLab: a \textsc{Matlab} Drone Swarm Simulator}

\author{
Enrica Soria, \textit{Student Member, IEEE}, Fabrizio Schiano, \textit{Member, IEEE}, \\ and Dario Floreano, \textit{Senior Member, IEEE},%
\thanks{The authors are with the \href{www.lis.epfl.ch}{Laboratory of Intelligent Systems (LIS)}, École Polytechnique Fédérale de Lausanne (EPFL), 1015 Lausanne, Switzerland. }
\thanks{Corresponding author: {\tt\footnotesize \href{mailto:enrica.soria@epfl.ch}{enrica.soria@epfl.ch}}}
}

\newcommand{\mbb}[1]{\mathbb{#1}}

\newcommand{\norm}[1]{\|#1\|}

\newcommand{\Real}{\mbb{R}}

\def\baselinestretch{0.997}

\input{notation}

\begin{document}

\maketitle
\thispagestyle{empty}
\pagestyle{empty}


\begin{abstract}
Among the available solutions for drone swarm simulations, we identified a lack of simulation
frameworks that allow easy algorithms prototyping, tuning, debugging and performance analysis. Moreover, users who want to dive in the research field of drone swarms often need to interface with multiple programming languages. We present SwarmLab, a software entirely written in \textsc{Matlab}, that aims at the creation of standardized processes and metrics to quantify the performance and robustness of swarm algorithms, and in particular, it focuses on drones.
We showcase the functionalities of SwarmLab by comparing two decentralized algorithms from the state of the art for the navigation of aerial swarms in cluttered environments, Olfati-Saber's and Vasarhelyi's. We analyze the variability of the inter-agent distances and agents' speeds during flight. We also study some of the performance metrics presented, i.e. order, inter- and extra-agent safety, union, and connectivity. While Olfati-Saber's approach results in a faster crossing of the obstacle field, Vasarhelyi's approach allows the agents to fly smoother trajectories, without oscillations.
We believe that SwarmLab is relevant for both the biological and robotics research communities, and for education, since it allows fast algorithm development, the automatic collection of simulated data, the systematic analysis of swarming behaviors with performance metrics inherited from the state of the art.

\vspace{5pt}
\end{abstract}





\section*{Supplementary Material}\label{sec:supplementary-material}

\noindent 
Supplementary video: \href{https://youtu.be/xMXA9OWSxe8}{https://youtu.be/xMXA9OWSxe8}.\\
SwarmLab is available on Github: \href{https://github.com/lis-epfl/swarmlab}{https://github.com/lis-epfl/swarmlab}.

\section{Introduction}\label{sec:introduction}

{Drone} applications have exploded in the last decade
and, recently, the availability of inexpensive hardware has awoken the interest for aerial swarms where several flying robots collaborate to achieve a collective task~\cite{floreano_science_2015,schilling2019learning}. 
Benefits of multi-drone systems are envisioned for a wide range of missions including search and rescue~\cite{bernard2011autonomous}, long-term monitoring~\cite{zhang2016seeing}, sensor data collection~\cite{erman2008enabling}, indoor navigation~\cite{stirling_indoor_2012}, environment exploration~\cite{mcguire_minimal_2019}, and cooperative grasping and transportation~\cite{mellinger2013cooperative}.
In the entertainment industry, the latest challenge is the coordination of fleets of hundreds of drones that light up the night sky with aerial shows, as Intel\footnote{\href{https://intel.com/content/www/us/en/technology-innovation/aerial-technology-light-show.html}{https://intel.com/content/www/us/en/technology-innovation/aerial-technology-light-show.html}} and Ehang\footnote{\href{https://ehang.com/formation}{https://ehang.com/formation}} have displayed.

\begin{figure}[t!]
    \centering
    \includegraphics[width=\columnwidth]{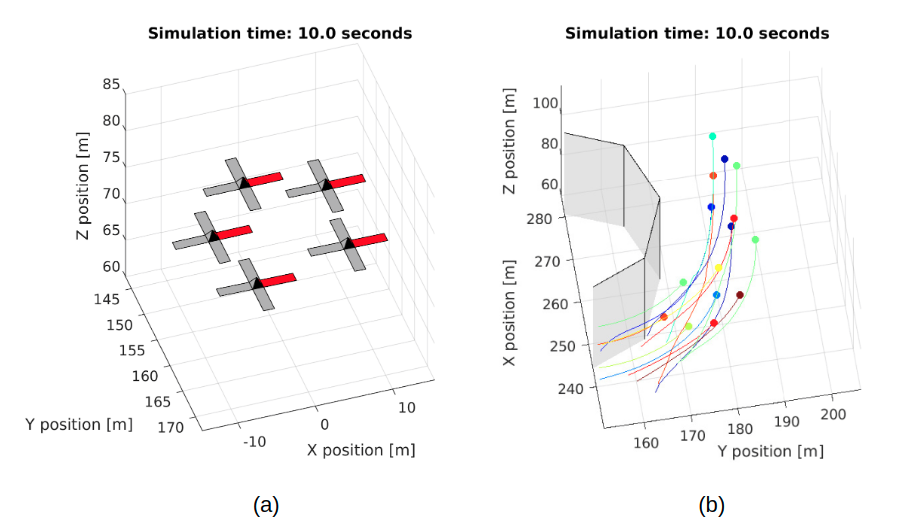}
    \vspace{-1em}
    \caption{%
        \textbf{3D swarm visualizations in SwarmLab}. In (a), 5 quadcopter drones coordinate in collective flight, while, in (b), a swarm of 15 drones simulated with point-mass dynamics executes a collision avoidance maneuver around an obstacle. Both snapshots are captured at $ 10 s $ of simulation.
    }\label{fig:overview}
    \vspace{-1.3em}
\end{figure}

However, the first step towards the deployment of such complex systems in real-world scenarios is simulation\cite{andrews2012simulation}.
The development of algorithms and applications for autonomous aerial vehicles requires the availability of a suitable simulation framework for rapid prototyping and simulation in reproducible scenarios. This is desirable in all robotics fields, but it is especially relevant for collective systems such as drone swarms, where errors can propagate through the individuals and lead to catastrophic results~\cite{huang_understanding_2013}. Although multiple open-source frameworks exist for simulating aerial robots~\cite{2012simpar_meyer,michel2004cyberbotics,rohmer_v-rep_2013,shah_airsim_2018}, the majority are focused on the realism of a single robot and cannot manage a large number of drones in real-time. On the other side, simulators that support multiple robots do not implement the nonlinear robot's dynamics or they require the interaction with several programming languages. Besides, there is no framework that provides ready-to-use control algorithms, debugging tools and performance analysis functionalities for aerial swarms. The potential user has to develop their own tools compatible with the chosen framework, which are not standard and prone to error.

\begin{figure}[h!]
    \centering
    \includegraphics[width=\columnwidth]{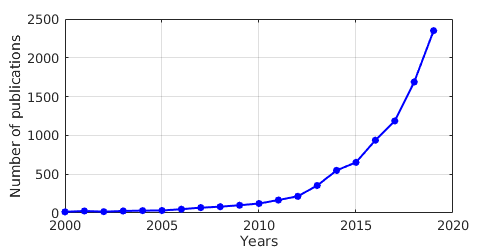}
    \vspace{-1em}
    \caption{%
        \textbf{\textit{Aerial swarms} publications}. Number of publications containing the words \textit{`aerial swarms`} between the years 2000 and 2019. Source:~\href{http://apps.webofknowledge.com}{http://apps.webofknowledge.com}.
    }\label{fig:publications}
    \vspace{-1em}
\end{figure}

In this work we propose SwarmLab, a simulator for single drones and drone swarms. Its main goal is to propose an alternative to existing robotic simulation solutions, that is explicitly centered on drone swarms (see Fig.~\ref{fig:overview}). 
We used \textsc{Matlab}\footnote{\href{https://mathworks.com/products/matlab.html}{https://mathworks.com/products/matlab.html}}, a scripting programming platform that allows us to implement and debug algorithms rapidly, use a large database of built-in functionalities, and create plots and videos with minimum effort. SwarmLab allows both accurate simulation of one drone, and efficient simulation of swarming behaviors with hundreds of agents. Concerning swarming, which is our focus, we provide support for fast instantiation of drone swarms, and the creation of environments with obstacles. We also include control algorithms from the state of the art, extensive plotting and debugging tools, 3D visualization functionalities, and performance analysis tools.
These features make our software relevant for the booming biological and robotics research communities in the field of aerial swarms (see Fig.~\ref{fig:publications}), and for education.

The rest of the paper is organized as follows. Section~\ref{sec:related-work} discusses available alternatives for robotic and, more specifically, drone simulations, while highlighting the gap that we aim to fill with SwarmLab. Section~\ref{sec:architecture} describes the architecture of the software and guides potential users through its main functionalities. Finally, Section~\ref{sec:use-case} shows how SwarmLab can be used for a comparative analysis of swarm algorithms and analyzes the computational time required for different simulation configurations.


\section{Related Work}\label{sec:related-work}

Currently, many robotic simulators are available on the market. A subset of them allows drone swarm simulations, but still require a considerable amount of time and programming languages for the prototyping and testing of aerial swarms algorithms, with the inconveniences stated in Sec~\ref{sec:introduction}.
A complete survey of the state of the art in robotic simulation is beyond the scope of this paper (see~\cite{pitonakova2018,staranowicz2011survey,de2019analysis} for a more detailed overview and performance comparison).
Instead, we aim to highlight the main features they offer and point out the needs that led to the development of SwarmLab.

Among the most well-known open-source 3D simulators for robots, we find Gazebo, WeBots, V-REP, ARGoS, and, more recently, AirSim ~\cite{koenig_design_2004,2012simpar_meyer,michel2004cyberbotics,rohmer_v-rep_2013,pinciroli_argos_2012,shah_airsim_2018}.
Gazebo~\cite{koenig_design_2004,2012simpar_meyer} can simulate the physics and dynamics of any mechanical structure modeled with joints, and it offers a large library of ready-to-use models, drones included.
Gazebo also allows integration with flight controller stacks for Software-In-The-Loop (SITL) drone simulation~\cite{meier_px4_2015}. Alternatively, RotorS is an extension to Gazebo designed for multirotors that includes example controllers, besides additional models and simulated sensors~\cite{Furrer2016,mccord2019distributed}. Users can code their functionalities in C++ and interface through ROS.
WeBots, recently released open-source, uses the same physics engine of Gazebo, but provides APIs for a large number of programming languages and includes drone models~\cite{michel2004cyberbotics} {natively}.
The V-REP simulator, now continued under the name of CoppeliaSim, offers features for easier editing of robots and other models~\cite{rohmer_v-rep_2013}. Development can be performed by means of the built-in Lua interpreter or by using a C or Python API.
More oriented to swarm robotics, ARGoS represents a lightweight alternative that offers a good tradeoff between scalability and extensibility~\cite{pinciroli_argos_2012}. It allows the user to simulate a larger number of robots {and it provides the possibility to use physics engines of different types, but it does not include drone models natively}. 
Robots can be programmed either through Lua scripts or in C++.
Specifically dedicated to drones and cars, AirSim~\cite{shah_airsim_2018} is a more recent simulator built on Unreal Engine~{\footnote{\href{https://www.unrealengine.com}{https://www.unrealengine.com}}} and as Gazebo, it allows SITL integration of flight controllers such as PX4. In AirSim, multi-agent simulations are easy to set up, and custom functionalities can be coded thanks to C++ and Python APIs.

All the simulators above are based on powerful 3D rendering engines, and they are mainly coded in C++. As a consequence, they provide graphical realism. However, they often require familiarity with more than one programming language. Also, they necessitate the addition of custom features for simulating an aerial swarm, which makes these simulators unsuitable for quick tests.
V-REP and WeBots include drone models natively, but drone swarm control and navigation algorithms must be designed, coded, and tuned by the user.

The simulators mentioned above are general-purpose robotic simulators. Instead, specific to drone simulation, we find the work by Beard et al.~\cite{beard_small_2012}. They describe fixed-wing drone systems with a waterfall architecture, where high-level blocks steer the drone to a goal destination, and lower-level blocks simulate physics and sensors. Driven by educational purposes, the authors released open-source templates in \textsc{Matlab} and Simulink\footnote{\href{https://github.com/randybeard/mavsim_template_files}{https://github.com/randybeard/mavsim\_template\_files}}. However, this work does not include quadcopter dynamics and swarming functionalities. However, it constitutes the foundation of SwarmLab. 

To the best of our knowledge, the only publicly available simulators geared towards aerial swarms are robotsim\footnote{\href{https://github.com/csviragh/robotsim}{https://github.com/csviragh/robotsim}}~\cite{vasarhelyi_optimized_2018} and the work of D'Urso et al.~\cite{durso_integrated_2019}. The first is a simulator written in C and, 
although it goes in the direction of SwarmLab,
no drone dynamics are implemented and architectural modularity is missing. The second is a software middleware that coordinates available tools (Gazebo, ArduCopter\footnote{\href{https://ardupilot.org/copter/}{https://ardupilot.org/copter/}} and ns-3\footnote{\href{https://www.nsnam.org/}{https://www.nsnam.org/}}) for the realistic simulation of the physics, graphics, flight control stack, and communication of interconnected drones and computers. This software is thought as a bridge towards a real-world implementation. However, its realism comes at the expense of its ease of use. Indeed, this simulator does not have the advantage of being contained within a single software such as \textsc{Matlab}.
Moreover, none of the simulators mentioned in this section provide functionalities for plotting, analysis, and performance assessment of the collective motion, which represents a limitation that we intend to overcome.



\section{Software architecture}\label{sec:architecture}

\begin{figure*}[!htb]
    \vspace{-1em}
    \centering
    \includegraphics[width=\textwidth]{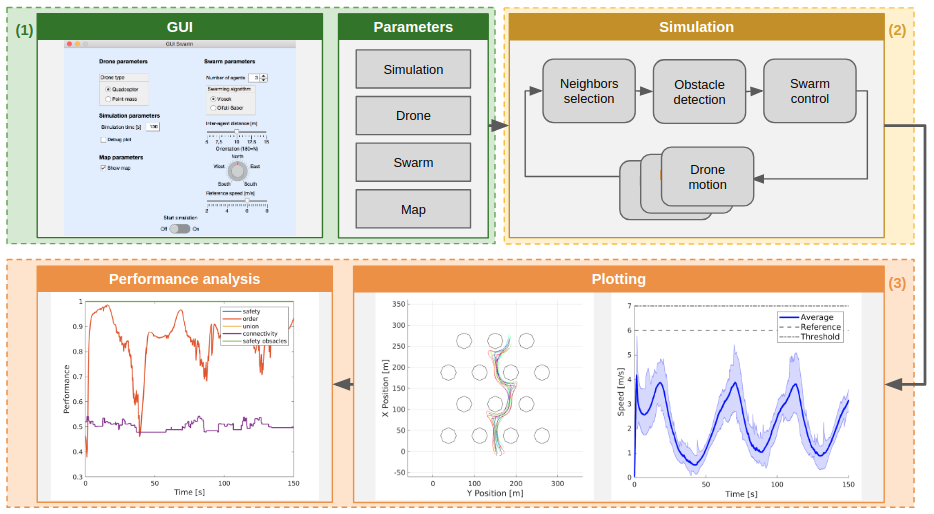}
    \vspace{-1em}
    \caption{%
        \textbf{SwarmLab simulation workflow}. From the top left, in clockwise order: (1) in the GUI, the user sets the parameters related to the simulation, drone typology, swarm algorithm and environment. Alternatively, parameters can be set in specific \textsc{Matlab} scripts. Then, he launches the simulation; (2) the main simulation loop computes control commands for the drones, based on the information of the map and neighboring drones; 
        (3) both real-time and post-simulation plotting {of the state variables} help the user with the analysis and debugging of the swarming behavior. {Moreover, at the end of the simulation the user can inspect the swarm performance metrics}.
    }\label{fig:architecture}
    \vspace{-0.8em}
\end{figure*}

SwarmLab is written in \textsc{Matlab} for several reasons. Firstly, this is a scripting language that operates at a high level of abstraction and therefore does not require extensive programming experience. Secondly, this framework provides several built-in toolboxes for design, control, analysis, and visualization of the studied systems, that reduce even further the programming effort and make it widely popular among the scientific community for education and research applications. {Moreover, code generation features are available to automatically translate the code to C/C++ and reduce the computational time or embed the algorithms on the robots controllers.}
SwarmLab follows the Object Oriented Programming (OOP) paradigm and its modular structure is made of the following main components:
\begin{itemize}
    \item parameter scripts for the single drone, swarm, and environment definitions;
    \item the \textit{Drone} and \textit{Swarm} classes;
    \item graphical classes that allow run-time and offline 3D visualization of the \textit{Drone}/\textit{Swarm}, their state variables and performance;
    \item example scripts and a README file that guide the user through the main functionalities of the simulator.
\end{itemize}



In this work, we call a swarm a set of $ N $ agents labeled by $ i \in \{1, \ldots, N \} $. The position and velocity of the $ i $-th agent in the inertial frame are denoted by $ \textbf{p}_i $ and $ \textbf{v}_i \in \Real^{3} $, respectively. To keep our notation concise we let $ d_{ij} = \norm{ \textbf{p}_j- \textbf{p}_i} $ represent the distance between two agents $ i $ and $ j $, where $ \norm{\cdot} $ denotes the Euclidean norm. 
We model the swarm with a directed sensing graph $\graph=(\calV,\,\calE)$, where the vertex set $\calV=\{1\ldots N \}$ represents the agents, and the edge set $\calE\subseteq \calV\times \calV$ contains the pairs of agents $(i,j)\in \calE$ for which agent $i$ can sense agent $j$. We denote as $\calN_i=\{j\in\calV|\,(i,\,j)\in\calE\}\subset \calV$ the set of neighbors of an agent $i$ in $\graph$.
Another concept we borrow from algebraic graph theory is the algebraic connectivity that can be measured through the so-called \textit{connectivity eigenvalue}. This is the second smallest eigenvalue of the Laplacian matrix~\cite{fiedler1989laplacian} associated with the undirected graph $\graph'$ obtained from $\graph$ and it is usually denoted by $\eig_2$. 
The agents are drones, and we will mainly  consider quadcopters. Also, we consider a set of  $ M $ obstacles labeled by $ m \in \{1, \ldots, M \} $ that populate the environment.

\subsection{Drone}\label{sec:drone-models}

The \textit{Drone} class represents the building block for simulating a swarm.
This class supports the definition of quadcopters or fixed-wing drones, {based on the models in~\cite{bouabdallah2007full} and ~\cite{beard_small_2012}} respectively. A \textit{Drone} instance is defined by:
\begin{itemize}
    \item parameters related to the chosen platform (e.g., mass, aerodynamic and control parameters),
    \item current state vector: ($ p_n, p_e, p_d , u, v, w, \phi, \theta, \psi, p, q, r ) \in \Real^{12} $. This vector is respectively composed by the north, east and down position coordinates in the inertial frame, the linear velocity measured along the $x, y, z$ axes of the body frame, three Euler angles describing the drone orientation, i.e. roll, pitch, and yaw, and the angular velocities measured in the body frame,
    \item path planning variables, including a list of waypoints,
    \item graphic variables for the visualization of the drone and the plotting of the state variables.
\end{itemize}
The methods provided by this class allow the creation of new instances, the computation of the kinematics and dynamics based on the physical parameters, and the control of the drone thanks to one of the two autopilots tailored for either quadcopters or fixed-wing drones. Moreover, for the simulation of a single-drone mission, high-level functionalities for path navigation are provided, following the same structure of~\cite{beard_small_2012}.

\subsection{Swarm}\label{sec:swarm-models}

The \textit{Swarm} class contains the necessary properties and methods to instance, initialize and manage \textit{Swarm} objects. These objects are made of an ensemble of dynamic agents of type \textit{Drone}. Their fundamental {properties} are:
\begin{itemize}
    \item \textit{drones}: a vector of \textit{Drone} objects,
    \item \textit{nb\_agents}: the number of agents included in the swarm,
    \item \textit{algorithm}: the selected algorithm for swarm navigation.
\end{itemize}

The workflow of a swarm simulation is summarized in Fig.~\ref{fig:architecture}. The user can start a simulation either by running an example script or by interacting with a Graphical User Interface (GUI) that accounts for real-time changes of the swarm parameters. When the simulation starts, 
a number of \textit{Drone} instances are created and added to the \textit{Swarm}. Also, the user can decide to instance a \textit{swarm viewer} to visualize the evolution of the swarm state during the simulation time. 
The main simulation loop computes at every iteration the control commands for every drone of the swarm and updates their states. Control commands for a given drone $i$ only depend on its neighbors $ \calN_i $.
Depending on the user's choice, the \textit{Swarm} class uses a different swarm algorithm to compute commands for every drone. Alternatively, the user can implement and test their own control algorithm as a method of the \textit{Swarm} class where the drones' states are accessible, by following the available examples.

\begin{figure}[t!]
    \vspace{-1em}
    \centering
    \includegraphics[width=\columnwidth]{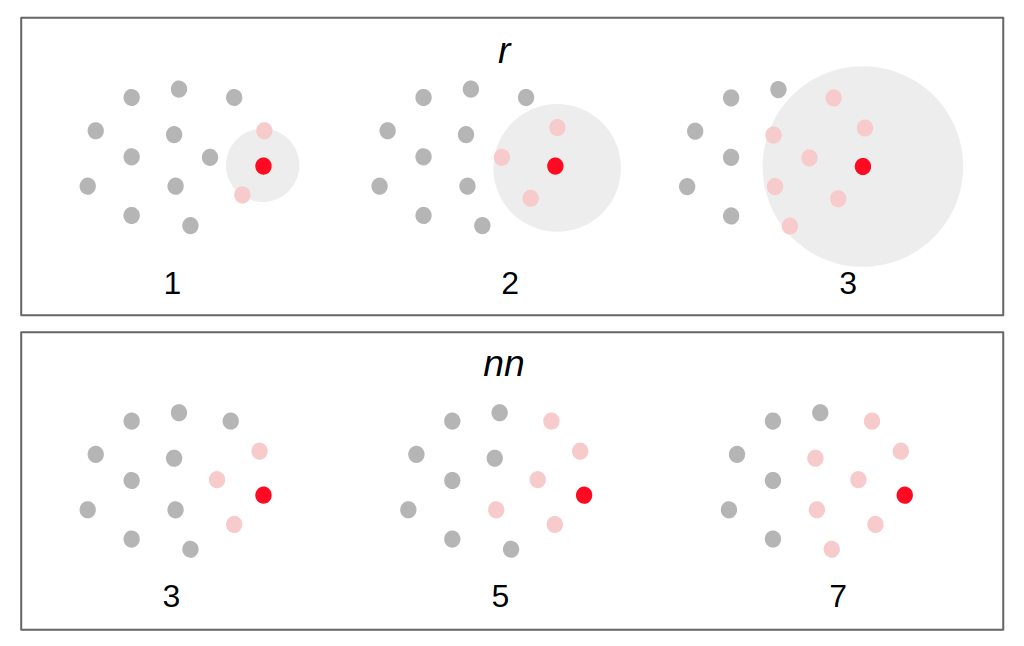}
    \vspace{-1em}
    \caption{%
        \textbf{Neighbor selection}. Illustration of different {neighborhoods} for varying values of the euclidean (first row) and topological distance (second row). The focal agent $i$, i.e. the agent taken into account, is highlighted in red, while its neighbors $ j \in \calN_i $ are in light red. On top, the shaded areas indicate the influence area of the focal agent.
    }\label{fig:neig-selection}
    \vspace{-0.8em}
\end{figure}

\subsection{Swarm algorithms}\label{sec:swarming-algorithms}
In SwarmLab, we implemented and adapted two {representative algorithms belonging to the category of decentralized swarming. The reason for this choice is that a decentralized approach can make the system easily scalable and robust to the failures of a single individual.} The first algorithm is authored by Olfati-Saber, who proposes a formal theoretical framework for the design and analysis of swarm algorithms based on potential fields and graph theory~\cite{olfati-saber_flocking_2006}. It is based on the construction of a \textit{collective potential} that penalizes the deviation of the agents from a lattice shape. In addition, a \textit{consensus} term makes the agents agree on their speed and velocity direction. At the equilibrium, in the absence of obstacles, the agents occupy positions at a constant distance from their neighbors and translate with constant velocity.
The second algorithm we implemented is an adaptation of the recent Vasarhelyi's algorithm, defined by the following rules: \textit{repulsion} to avoid inter-agent collisions, \textit{velocity alignment} to steer the agents to an average direction, and \textit{self-propulsion} to match a preferred speed value~\cite{vasarhelyi_optimized_2018}. In addition, the algorithm includes friction forces that reduce oscillations and ease the implementation on real robots.
Finally, both algorithms propose an \textit{obstacle avoidance} behavior to deal with convex obstacles.
{In several engineering applications (e.g., mapping, area coverage, search and rescue), we require the swarm to fly in a specific direction. To this aim, we allow the selection between the consensus on velocity in Olfati-Saber's algorithm, or the velocity alignment in Vasarhelyi's algorithm and a so-called \textit{migration} term that penalizes deviations from a given velocity.}

{In decentralized approaches, one agent's movement is only influenced by local information coming from its neighbors.}
Neighbors selection can be operated according to different metrics. Two widely adopted ones are the euclidean and the topological distances~\cite{strandburg-peshkin_visual_2013, ballerini_interaction_2008}. The euclidean distance defines $ \calN_i $ as the set of agents $ j \ne i $ within a constant radius of influence $ r $ from agent $ i $. The cardinality of this set depends on the density of the swarm. Instead, the topological distance defines $ \calN_i $ as the number $ nn $ of nearest agents to $ i $, as illustrated in Fig.~\ref{fig:neig-selection}. In the latter case, the cardinality does not depend on the density. In our software, both distances are implemented, and they can be set before starting the simulation. 

\begin{figure}[t!]
    \vspace{-1em}
    \centering
    \includegraphics[width=\columnwidth]{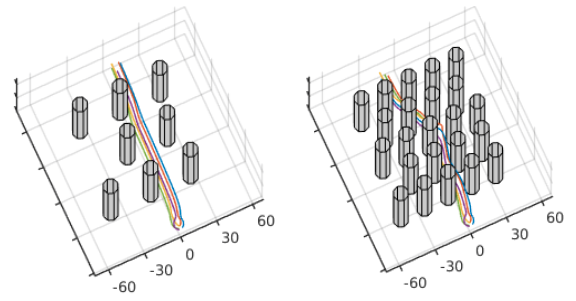}
    \vspace{-1em}
    \caption{%
        \textbf{Maps with varying obstacle density}. Cylindrical obstacles are distributed on the map to reproduce a forest-like environment. The obstacle density increases from left to right. Thanks to the obstacle avoidance behavior, the swarm agents are able to avoid collisions with the environment.
    }\label{fig:environments}
    \vspace{-0.8em}
\end{figure}

For the navigation in cluttered environments, we provide a \textit{map} that generates cylindrical obstacles with parametric size and density (see Fig.~\ref{fig:environments}).
In both Olfati-Saber's and Vasarhelyi's algorithms, the obstacle avoidance behavior is modeled via virtual agents. These are additional agents to which we assign a position and velocity that depend on the obstacles configuration, and they act on drones as if they were normal agents.

SwarmLab offers two modalities for swarm simulation: the high-fidelity mode simulates quadcopter drones, where realistic dynamics and control are implemented, while the second approximates the drone dynamics with the dynamics of a point mass, whose state is defined by inertial position and velocity. This is meant to trade-off simulation fidelity and computational efficiency (see Sec.~\ref{sec:conclusions} for more details).


\subsection{Graphical User Interfaces (GUIs)}\label{sec:gui}

For introducing the user to the simulator functionalities we provide two GUIs: one for selecting the parameters related to single drones simulations and one for aerial swarms simulations. The latter is split into sections that allow the user to select the drone dynamics, either quadcopter or point-mass, the main swarm parameters such as the number of drones, the preferred value of the inter-agent distance, the speed and orientation of the swarm motion, and simulation parameters such as the simulation time duration, the presence of debugging plots and the creation of a map with obstacles.

\subsection{Plotting tools}\label{sec:plotting}

One of the most critical parts of programming is verifying the validity of the code and algorithms. To this aim, a user needs tools to analyze the state of the system and find the origin of potential faults. 
SwarmLab allows the tracking of: (i) inter-agent distance and distance to obstacles, in order to detect collisions, (ii) swarm speed, useful for instance to monitor slow-down effects in front of obstacles, (iii) acceleration, to observe its variability and, hence, the efficiency of the algorithm. State plotting is possible both during the simulation, \textit{run-time}, and at the end, \textit{offline}. Run-time is useful for debugging, while offline is practical when the user does not want to slow down the simulation with the addition of graphic features. 
{Single-drone plotting can be used simultaneously to observe the state of a specific drone in the swarm.}

\subsection{Performance analysis}\label{sec:performance}

The presence of obstacles in the environment can threaten the ability of the agents to remain cohesive during their mission and prevent them from flying smoothly in the migration direction. 
In these situations, the swarm may split into multiple subgroups with no influence on one another, and collisions may occur. To evaluate the collective navigation performance during flight, we use five metrics adopted in previous work~\cite{soria_influence_2019}. These metrics were inspired by robotic and biological studies of aerial swarms:
\begin{itemize} 
  \item the \textit{order} metric, $\Phi_{\text{o}}$: it captures the correlation of the agents' movements and gives an indication about how ordered the flock is. We express it by 
  \begin{equation*}
  \vspace{-0.5em}
  \Phi_{\text{o}} = \frac{1}{N(N-1)} \sum\limits_{i,j\neq i} \frac{{\bm{v}}_i \cdot {\bm{v}}_j}
  {\norm{{\bm{v}}_i} \norm{{\bm{v}}_j}}.
  \vspace{-0em}
  \end{equation*}
  \item The \textit{safety} metrics, $\Phi_{\text{s,ag}}$ and $\Phi_{\text{s,obs}}$: they respectively measure the risk of collisions among the swarm agents or between agents and obstacles. We denote with $ r_{\text{ag}} $ the collision radius of an  agent. Instead, $ r_{obs} $ denotes the obstacle radius. The number of inter-agent collisions is $n_{\text{ag}} = |\{(i,j) \ s.t.\ j \neq i\ \land d_{ij}< 2 r_{\text{ag}} \}|$, while the number of collisions with obstacles is $n_{\text{obs}} = |\{(i,m) \ \land d_{im}< r_{ag} + r_{\text{obs}} \}|$. Therefore, the inter-agent safety and the safety with obstacles can be expressed as 
  \begin{equation*}
  \vspace{-0.5em}
    \Phi_{\text{s,ag}}=1-\frac{n_{ag}}{N(N-1)} \ \ \ , \ \ \ \Phi_{\text{s,obs}}=1-\frac{n_{obs}}{N}.
    \vspace{-0em}
  \end{equation*}
  \item The \textit{union} metric, $\Phi_{\text{u}}$: it counts the number of independent subgroups that originates during the simulation. We define $ n_{c} $ as the number of connected components of the undirected graph that corresponds to the flock topology, then it holds 
  \begin{equation*}
  \vspace{-0.5em}
    \Phi_{\text{u}} = 1 - \frac{n_{c}-1}{N-1}.
    \vspace{-0em}
  \end{equation*}
  \item The \textit{connectivity} metric, $\Phi_{\text{c}}$: it is defined from the algebraic connectivity of the sensing graph that underlines the considered swarm configuration as 
  \begin{equation*}
  \vspace{-0.5em}
    \Phi_{\text{c}} =\frac{\lambda_2}{N}
    \vspace{-0em}
  \end{equation*}
  where $\lambda_2$ is defined above. Notice that $\Phi_{\text{c}} \neq 0$ only when $\Phi_{\text{u}}=1$. In this sense, the connectivity metric is complementary to the union metric.

%
\end{itemize}



\begin{figure*}[t]
    \includegraphics[width=\textwidth]{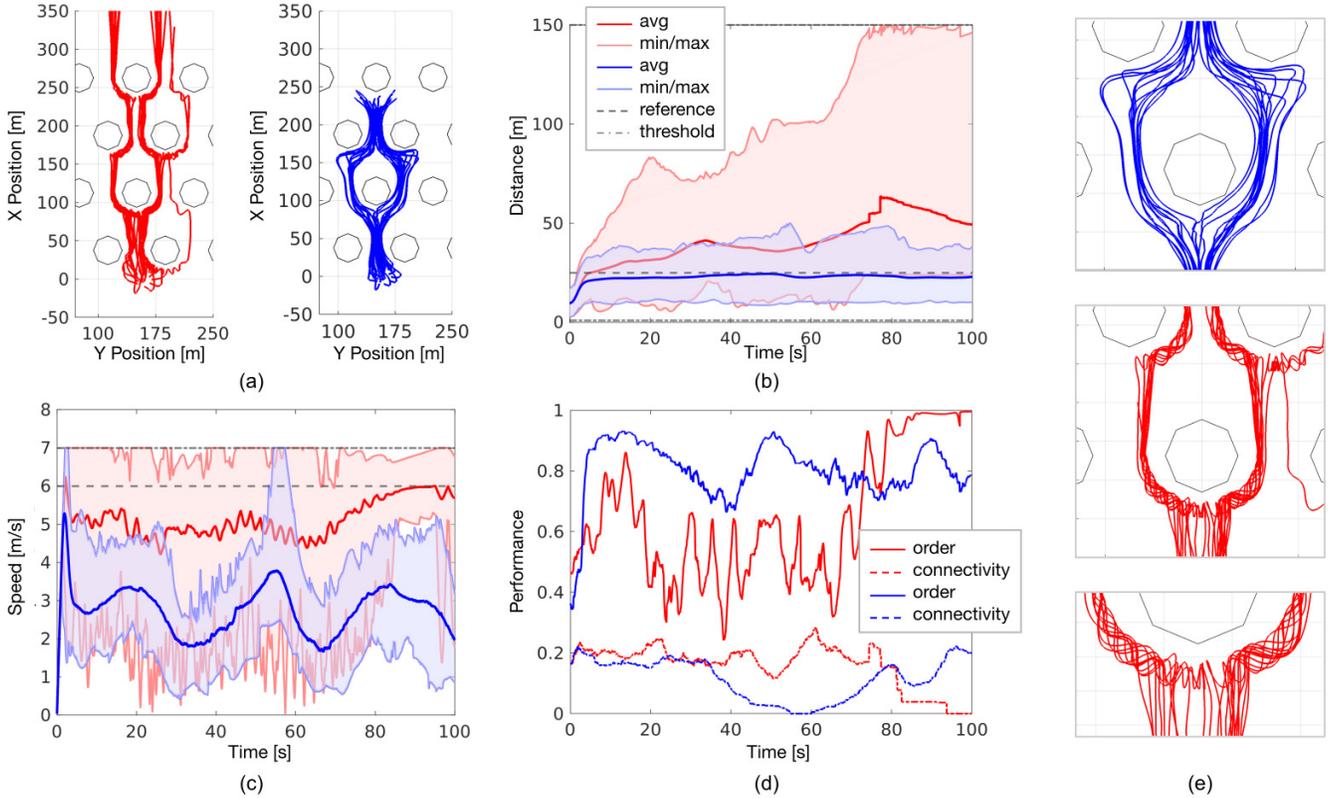}
    \vspace{-1em}
    \caption{\textbf{Comparison of two swarm algorithms}. Olfati-Saber's plots are in red, while Vasarhelyi's plots are in blue. The simulation time is $ 100s $ for both algorithms. In (a) we observe the top view of the trajectories of 25 agents, flying in an obstacle field from lower to higher values of the x position. In (b) and (c) inter-agent distances and speeds are compared, in terms of average, minimum, and maximum values. The reference values are in dashed lines, while the collision threshold, the radius of influence $ r $, and the maximum speed are in dashdotted lines. Two of the presented performance metrics are compared in (d), the order ($\Phi_{\text{o}}$) and the connectivity ($\Phi_{\text{c}}$). Finally, (e) shows the zoom in of the trajectories of the agents around obstacles, from above.}\label{fig:comparison}
    \vspace{-1em}
\end{figure*}

\section{{Comparison of swarm algorithms and computational time analysis}}\label{sec:use-case}

{In this section, we present the results of the comparison of two swarm algorithms enabled by SwarmLab. Moreover, we present an analysis of the computational time of the simulator in different modes.}

To compare swarm algorithms, we present a use case where 25 agents fly in a cluttered environment. We select point-mass dynamics, and  we perform the neighbor selection with $ nn = 10 $ and $ r = 150 $. The agents' initial positions are randomly selected in a cubic volume, and the swarm is let navigate over $ 100 s $ in the direction of increasing values of the x position. Both swarm algorithms described in Sec.~\ref{sec:swarm-models} are tested and the graphical outputs are reported in Fig.~\ref{fig:comparison}. We notice that Olfati-Saber's algorithm prioritizes the tracking of the speed reference value (see Fig.~\ref{fig:comparison}c), while Vasarhelyi's one allows the agent to slow down in front of obstacles to better match the reference inter-agent distance (see Fig.~\ref{fig:comparison}b). The minimum distance threshold in Fig.~\ref{fig:comparison}b is never crossed with both algorithms, which means that no inter-agent collisions occur. By examining the trajectories, in Fig.~\ref{fig:comparison}a and Fig.~\ref{fig:comparison}e, we see that the obstacle avoidance behavior of the second algorithm allows a smoother interaction of the agents with obstacles and reduces their oscillations, while in the first case, both in the trajectories and speed we observe prominent oscillations. Concerning the order $\Phi_{\text{o}}$, better performance is obtained with Vasarhelyi's algorithm (see Fig.~\ref{fig:comparison}d). Indeed, oscillations around obstacles prevent ordered flight in the case of Olfati-Saber's swarming. On the contrary, at the end of the simulation, Olfati-Saber's swarm order is higher. Indeed, once that the agents quit the obstacle field, in the free space, their velocity converges to the migration one. Contrarily, while the agents fly among obstacles, connectivity $\Phi_{\text{c}}$ is slightly better in Olfati-Saber's case, and vice versa in the free space. {Being connectivity related to the speed of the information flow among the agents, good values are preferred in scenarios where information-sharing among the agents is crucial (e.g., cooperative localization).}

\begin{figure}[h!]
    \vspace{-0.8em}
    \centering
    \includegraphics[width=\columnwidth]{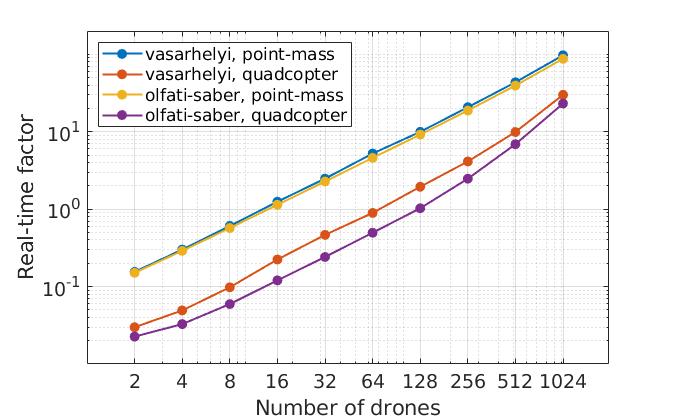}
    \vspace{-1em}
    \caption{%
    \textbf{Real-time factor for varying sizes of the swarm}. The number of drones goes from 2 to 1024. Two swarm algorithms (Vasarhelyi's and Olfati-Saber's) and two dynamics (point-mass and quadcopter) are compared.
    }\label{fig:computational-time}
    \vspace{-1em}
\end{figure}

To evaluate the computational time, we run the two swarm algorithms with up to 1024 agents without any graphical output. The simulation time is arbitrarily set to 20 seconds. The hardware used is a DELL Precision Tower with a 3.6 GHz Intel Core i7-7700 processor and 16 GB 2400 MHz RAM.
The results are reported in Fig.~\ref{fig:computational-time}, where the computational time is expressed in terms of \textit{real-time factor}. A \textit{real-time factor} equal to one means that the computational time required by the computer to run the simulation is equal to the simulation time. Instead, a value equal to two indicates that the computational time is twice the simulation time. 
The trend we notice is the same for both swarm algorithms and both drone typologies. As expected, when modelling the drones as point-masses the real-time factor is significantly lower. For instance, when we consider a swarm of 64 agents the real-time factor is close to 0.5 for the Vasarhelyi's algorithm and 0.9 for the Olfati-Saber's algorithm. Instead, when a full nonlinear quadcopter dynamics is used with the same amount of agents, the real-time factor increases up to 4.6 for the Vasarhelyi's algorithm and 5.2 for the Olfati-Saber's one. 

\section{Conclusions and future work}\label{sec:conclusions}

In this paper, we presented a versatile and scalable drone swarm simulator entirely written in \textsc{Matlab} that integrates built-in functionalities for collective navigation, debugging of the algorithms, and performance analysis.
We believe that this framework can serve as a development tool and a comparative platform for the growing research community in aerial swarms, and for education. With reduced coding effort, the user can change parameters, edit their code, run and test it in a single scripted programming language.
Regarding future work, we will focus on the improvement of the computational time to allow faster simulation of large swarms. For this, we will consider automatic C/C++ code generation from \textsc{Matlab}.
Moreover, noise modeling and delays should be considered to narrow the gap between simulation and reality.
Finally, another challenge for future works is the integration of automatic parameter tuning for the swarm algorithms as done in~\cite{vasarhelyi_optimized_2018}. This will allow to optimize the swarming behavior for a given environment or task with respect to the implemented performance metrics.


\section*{Acknowledgments}\label{sec:acknowledgments}
We thank Andrea Giordano, Victor Delafontaine, Fabian Schilling, and Anthony De Bortoli for their valuable contribution. This work was supported by the Swiss National Science Foundation with grant
number 200020\_188457 and the European project RoboCom++. 

\bibliographystyle{IEEEtran}
\bibliography{bib.bib}

\end{document}

%% file: notation.tex
\newcommand{\kron}{\otimes}

\newcommand{\matr}[1]{\boldsymbol{#1}}

\newcommand{\zeros}[2]{
\ifthenelse{\equal{#2}{1}}{\vect{0}_{#1}}{\matr{\cancel{O}}_{#1 \times #2}}
}
\newcommand{\ones}[2]{
\ifthenelse{\equal{#2}{1}}{\vect{1}_{#1}}{\matr{1}_{#1 \kron #2}}
}

\newcommand{\graph}{\mathcal{G}}

\newcommand{\eig}{\lambda}

\newcounter{simulationcase}

\newcommand{\calV}{\mathcal{V}}
\newcommand{\calE}{\mathcal{E}}

\newcommand{\calN}{\mathcal{N}}

\newcommand{\mm}[1]{}  

\definecolor{mygray}{gray}{0.75} 
\newcommand{\old}[1]{}  





%% file: root.bbl
\ifdefined\DeclarePrefChars\DeclarePrefChars{'’-}\else\fi
\begin{thebibliography}{10}
\providecommand{\url}[1]{#1}
\csname url@samestyle\endcsname
\providecommand{\newblock}{\relax}
\providecommand{\bibinfo}[2]{#2}
\providecommand{\BIBentrySTDinterwordspacing}{\spaceskip=0pt\relax}
\providecommand{\BIBentryALTinterwordstretchfactor}{4}
\providecommand{\BIBentryALTinterwordspacing}{\spaceskip=\fontdimen2\font plus
\BIBentryALTinterwordstretchfactor\fontdimen3\font minus
  \fontdimen4\font\relax}
\providecommand{\BIBforeignlanguage}[2]{{%
\expandafter\ifx\csname l@#1\endcsname\relax
\typeout{** WARNING: IEEEtran.bst: No hyphenation pattern has been}%
\typeout{** loaded for the language `#1'. Using the pattern for}%
\typeout{** the default language instead.}%
\else
\language=\csname l@#1\endcsname
\fi
#2}}
\providecommand{\BIBdecl}{\relax}
\BIBdecl

\bibitem{floreano_science_2015}
D.~Floreano and R.~J. Wood, ``Science, technology and the future of small
  autonomous drones,'' \emph{Nature}, vol. 521, no. 7553, pp. 460--466.

\bibitem{schilling2019learning}
F.~Schilling, J.~Lecoeur, F.~Schiano, and D.~Floreano, ``Learning vision-based
  flight in drone swarms by imitation,'' \emph{IEEE Robotics and Automation
  Letters}, vol.~4, no.~4, pp. 4523--4530, 2019.

\bibitem{bernard2011autonomous}
M.~Bernard, K.~Kondak, I.~Maza, and A.~Ollero, ``Autonomous transportation and
  deployment with aerial robots for search and rescue missions,'' vol.~28,
  no.~6, pp. 914--931, 2011.

\bibitem{zhang2016seeing}
J.~Zhang, J.~Hu, J.~Lian, Z.~Fan, X.~Ouyang, and W.~Ye, ``Seeing the forest
  from drones: Testing the potential of lightweight drones as a tool for
  long-term forest monitoring,'' vol. 198, pp. 60--69, 2016.

\bibitem{erman2008enabling}
A.~T. Erman, L.~van Hoesel, P.~Havinga, and J.~Wu, ``Enabling mobility in
  heterogeneous wireless sensor networks cooperating with uavs for
  mission-critical management,'' vol.~15, no.~6, pp. 38--46, 2008.

\bibitem{stirling_indoor_2012}
T.~Stirling, J.~Roberts, J.-C. Zufferey, and D.~Floreano, ``Indoor navigation
  with a swarm of flying robots,'' in \emph{2012 {{IEEE International
  Conference}} on {{Robotics}} and {{Automation}}}.\hskip 1em plus 0.5em minus
  0.4em\relax {IEEE}, 2012, pp. 4641--4647.

\bibitem{mcguire_minimal_2019}
K.~N. McGuire, C.~D. Wagter, K.~Tuyls, H.~J. Kappen, and G.~C. H.~E. de~Croon,
  ``Minimal navigation solution for a swarm of tiny flying robots to explore an
  unknown environment,'' \emph{Science Robotics}, vol.~4, no.~35, p. eaaw9710,
  2019.

\bibitem{mellinger2013cooperative}
D.~Mellinger, M.~Shomin, N.~Michael, and V.~Kumar, ``Cooperative grasping and
  transport using multiple quadrotors,'' in \emph{Distributed autonomous
  robotic systems}, 2013, pp. 545--558.

\bibitem{andrews2012simulation}
P.~S. Andrews, S.~Stepney, and J.~Timmis, ``Simulation as a scientific
  instrument,'' in \emph{Proceedings of the 2012 workshop on complex systems
  modelling and simulation, Orleans, France}, 2012.

\bibitem{huang_understanding_2013}
B.~Huang, C.~Yu, and B.~D.~O. Anderson, ``Understanding {{Error Propagation}}
  in {{Multihop Sensor Network Localization}},'' vol.~60, no.~12, pp.
  5811--5819, 2013.

\bibitem{2012simpar_meyer}
J.~Meyer, A.~Sendobry, S.~Kohlbrecher, U.~Klingauf, and O.~von Stryk,
  ``Comprehensive simulation of quadrotor uavs using ros and gazebo,'' in
  \emph{3rd Int. Conf. on Simulation, Modeling and Programming for Autonomous
  Robots (SIMPAR)}, 2012.

\bibitem{michel2004cyberbotics}
O.~Michel, ``Cyberbotics ltd. webots™: professional mobile robot
  simulation,'' \emph{International Journal of Advanced Robotic Systems},
  vol.~1, no.~1, 2004.

\bibitem{rohmer_v-rep_2013}
E.~Rohmer, S.~P.~N. Singh, and M.~Freese, ``V-{{REP}}: {{A}} versatile and
  scalable robot simulation framework,'' in \emph{2013 {{IEEE}}/{{RSJ
  International Conference}} on {{Intelligent Robots}} and {{Systems}}}, 2013,
  pp. 1321--1326.

\bibitem{shah_airsim_2018}
S.~Shah, D.~Dey, C.~Lovett, and A.~Kapoor, ``{{AirSim}}: {{High}}-{{Fidelity
  Visual}} and {{Physical Simulation}} for {{Autonomous Vehicles}},'' in
  \emph{Field and {{Service Robotics}}}.\hskip 1em plus 0.5em minus 0.4em\relax
  {Springer International Publishing}, 2018, pp. 621--635.

\bibitem{pitonakova2018}
L.~Pitonakova, M.~Giuliani, A.~Pipe, and A.~Winfield, \emph{{Feature and
  Performance Comparison of the V-REP , Gazebo and ARGoS Robot
  Simulators}}.\hskip 1em plus 0.5em minus 0.4em\relax Springer International
  Publishing, 2018.

\bibitem{staranowicz2011survey}
A.~Staranowicz and G.~L. Mariottini, ``A survey and comparison of commercial
  and open-source robotic simulator software,'' in \emph{Proceedings of the 4th
  International Conference on PErvasive Technologies Related to Assistive
  Environments}, 2011.

\bibitem{de2019analysis}
M.~S.~P. de~Melo, J.~G. da~Silva~Neto, P.~J.~L. da~Silva, J.~M. X.~N. Teixeira,
  and V.~Teichrieb, ``Analysis and comparison of robotics 3d simulators,'' in
  \emph{2019 21st Symposium on Virtual and Augmented Reality (SVR)}.\hskip 1em
  plus 0.5em minus 0.4em\relax IEEE, 2019, pp. 242--251.

\bibitem{koenig_design_2004}
N.~Koenig and A.~Howard, ``Design and use paradigms for gazebo, an open-source
  multi-robot simulator,'' in \emph{2004 {{IEEE}}/{{RSJ International
  Conference}} on {{Intelligent Robots}} and {{Systems}} ({{IROS}}) ({{IEEE
  Cat}}. {{No}}.{{04CH37566}})}, vol.~3.\hskip 1em plus 0.5em minus 0.4em\relax
  {IEEE}, 2004, pp. 2149--2154.

\bibitem{pinciroli_argos_2012}
C.~Pinciroli, V.~Trianni, R.~O’Grady, G.~Pini, A.~Brutschy, M.~Brambilla,
  N.~Mathews, E.~Ferrante, G.~Di~Caro, F.~Ducatelle, M.~Birattari, L.~M.
  Gambardella, and M.~Dorigo, ``{{ARGoS}}: A modular, parallel, multi-engine
  simulator for multi-robot systems,'' \emph{Swarm Intelligence}, vol.~6,
  no.~4, pp. 271--295, 2012.

\bibitem{meier_px4_2015}
L.~Meier, D.~Honegger, and M.~Pollefeys, ``{{PX4}}: {{A}} node-based
  multithreaded open source robotics framework for deeply embedded platforms,''
  in \emph{2015 {{IEEE International Conference}} on {{Robotics}} and
  {{Automation}} ({{ICRA}})}.\hskip 1em plus 0.5em minus 0.4em\relax {IEEE},
  2015, pp. 6235--6240.

\bibitem{Furrer2016}
F.~Furrer, M.~Burri, M.~Achtelik, and R.~Siegwart, \emph{Robot Operating System
  (ROS): The Complete Reference (Volume 1)}.\hskip 1em plus 0.5em minus
  0.4em\relax Cham: Springer International Publishing, 2016, ch. RotorS---A
  Modular Gazebo MAV Simulator Framework, pp. 595--625.

\bibitem{mccord2019distributed}
C.~McCord, J.~P. Queralta, T.~N. Gia, and T.~Westerlund, ``Distributed
  progressive formation control for multi-agent systems: 2d and 3d deployment
  of uavs in ros/gazebo with rotors,'' in \emph{2019 European Conference on
  Mobile Robots (ECMR)}.\hskip 1em plus 0.5em minus 0.4em\relax IEEE, 2019.

\bibitem{beard_small_2012}
R.~W. Beard and T.~W. McLain, \emph{Small Unmanned Aircraft: Theory and
  Practice}.\hskip 1em plus 0.5em minus 0.4em\relax {Princeton University
  Press}, oCLC: ocn724663112.

\bibitem{vasarhelyi_optimized_2018}
G.~Vásárhelyi, C.~Virágh, G.~Somorjai, T.~Nepusz, A.~E. Eiben, and
  T.~Vicsek, ``Optimized flocking of autonomous drones in confined
  environments,'' \emph{Science Robotics}, vol.~3, no.~20, 2018.

\bibitem{durso_integrated_2019}
F.~D’Urso, C.~Santoro, and F.~F. Santoro, ``An integrated framework for the
  realistic simulation of multi-{{UAV}} applications,'' vol.~74, pp. 196--209.

\bibitem{fiedler1989laplacian}
M.~Fiedler, ``Laplacian of graphs and algebraic connectivity,'' \emph{Banach
  Center Publications}, vol.~25, no.~1, pp. 57--70, 1989.

\bibitem{bouabdallah2007full}
S.~Bouabdallah and R.~Siegwart, ``Full control of a quadrotor,'' in \emph{2007
  IEEE/RSJ International Conference on Intelligent Robots and Systems}.\hskip
  1em plus 0.5em minus 0.4em\relax Ieee, 2007, pp. 153--158.

\bibitem{olfati-saber_flocking_2006}
R.~Olfati-Saber, ``Flocking for {{Multi}}-{{Agent Dynamic Systems}}:
  {{Algorithms}} and {{Theory}},'' \emph{IEEE Transactions on Automatic
  Control}, vol.~51, no.~3, pp. 401--420, 2006.

\bibitem{strandburg-peshkin_visual_2013}
A.~Strandburg-Peshkin, C.~R. Twomey, N.~W.~F. Bode, A.~B. Kao, Y.~Katz, C.~C.
  Ioannou, S.~B. Rosenthal, C.~J. Torney, H.~S. Wu, S.~A. Levin, and I.~D.
  Couzin, ``Visual sensory networks and effective information transfer in
  animal groups,'' \emph{Current Biology}, vol.~23, no.~17, pp. R709--R711,
  00144.

\bibitem{ballerini_interaction_2008}
M.~Ballerini, N.~Cabibbo, R.~Candelier, A.~Cavagna, E.~Cisbani, I.~Giardina,
  V.~Lecomte, A.~Orlandi, G.~Parisi, A.~Procaccini, M.~Viale, and
  V.~Zdravkovic, ``Interaction ruling animal collective behavior depends on
  topological rather than metric distance: {{Evidence}} from a field study,''
  \emph{Proceedings of the National Academy of Sciences}, vol. 105, no.~4, pp.
  1232--1237, 2008.

\bibitem{soria_influence_2019}
E.~Soria, F.~Schiano, and D.~Floreano, ``The influence of limited visual
  sensing on the {{Reynolds}} flocking algorithm,'' \emph{{{IEEE Third
  International Conference}} on {{Robotic Computing}} ({{IRC}})}, 2019.

\end{thebibliography}
